%% file: root.tex
\documentclass[letterpaper, 10 pt, conference]{ieeeconf}  
\IEEEoverridecommandlockouts                              
\usepackage{hyperref}
\usepackage{url}
\usepackage{graphicx}
\usepackage{algorithm}
\usepackage{algorithmic}
\usepackage{graphics, times, amsfonts, amsmath, amssymb, cite, verbatim, comment}
\usepackage{tikz}
\usepackage{framed}
\usetikzlibrary{shapes,arrows}
\usepackage{pgfplots}
\usepackage{color}
\usepackage{flushend}
\usepackage{multirow}
\usepackage{rotating}
\usepackage{standalone}
\usepackage{pifont}
\usepackage[font=small]{caption}
\usepackage{subcaption}

\newtheorem{assumption}{\hspace{0pt}\bf Assumption}
\newtheorem{lemma}{\hspace{0pt}\bf Lemma}
\newtheorem{theorem}{\hspace{0pt}\bf Theorem}
\newtheorem{proposition}{\hspace{0pt}\bf Proposition}

\title{\LARGE \bf Doubly Random Parallel Stochastic Methods for Large Scale Learning}

\author{Aryan Mokhtari, Alec Koppel, and Alejandro Ribeiro
\thanks{Work in this paper is supported by ARO W911NF-10-1-0388, NSF CAREER CCF-0952867, and ONR N00014-12-1-0997. The authors are with the Department of Electrical and Systems Engineering, University of Pennsylvania, Philadelphia, PA 19143, USA
        {\tt\small \{aryanm, akoppel, aribeiro\}@seas.upenn.edu}}%
}

\input{mysymbol.sty}
\begin{document}

\maketitle

\begin{abstract}
We consider learning problems over training sets in which both, the number of training examples and the dimension of the feature vectors, are large. To solve these problems we propose the random parallel stochastic algorithm (RAPSA). We call the algorithm random parallel because it utilizes multiple processors to operate in a randomly chosen subset of blocks of the feature vector. We call the algorithm parallel stochastic because processors choose elements of the training set randomly and independently. Algorithms that are parallel in either of these dimensions exist, but RAPSA is the first attempt at a methodology that is parallel in both, the selection of blocks and the selection of elements of the training set. In RAPSA, processors utilize the randomly chosen functions to compute the stochastic gradient component associated with a randomly chosen block. The technical contribution of this paper is to show that this minimally coordinated algorithm converges to the optimal classifier when the training objective is convex. In particular, we show that: (i) When using decreasing stepsizes, RAPSA converges almost surely over the random choice of blocks and functions. (ii) When using constant stepsizes, convergence is to a neighborhood of optimality with a rate that is linear in expectation. RAPSA is numerically evaluated on the MNIST digit recognition  problem.
\end{abstract}

\input{Introduction.tex}

\input{Problem.tex}

\input{Convergence.tex}

\input{Simulations.tex}
\input{Conclusions.tex}

{
    \input{Appendix.tex}

\bibliographystyle{IEEEtran}
  \bibliography{bmc_article}
}

\end{document}

%% file: Introduction.tex

%
\section{Introduction}\label{sec_Introduction}

Learning is often formulated as an optimization problem that finds a classifier $\bbx^*\in\reals^p$ that minimizes the average of a loss function across the elements of a training set. For a precise definition consider a training set with $N$ elements and let $f_{n}:\reals^p\to\reals$ be a convex loss function associated with the $n$th element of the training set. The optimal classifier $\bbx^*\in\reals^p$ is defined as the minimizer of the average cost $F(\bbx) := (1/N)\sum_{n=1}^N f_{n}(\bbx)$,
\begin{equation}\label{eq:empirical_min}
   \bbx^* := \argmin_{\bbx} F(\bbx) 
          := \argmin_{\bbx}\frac{1}{N}\sum_{n=1}^N f_{n}(\bbx).
\end{equation}
Problems such as support vector machines, logistic regression, and matrix completion can be put in the form of problem \eqref{eq:empirical_min}. In this paper we are interested in large scale problems where both, the number of features $p$ and the number of elements $N$ in the training set are very large -- which arise, e.g., in text \cite{Sampson:1990:NLA:104905.104911}, image \cite{mairal2010online}, and genomic \cite{tacsan2014selecting} processing.

\begin{figure*}
\centering\input{diagram.tex}
 \caption{Random parallel stochastic algorithm (RAPSA). At each iteration, processor $P_i$ picks a random block from the set $\{\bbx_1,\dots,\bbx_B\}$ and a random set of functions from the training set $\{f_1,\dots,f_N\}$. The functions drawn are used to evaluate a stochastic gradient component associated with the chosen block. RAPSA is shown here to converge to the optimal argument $\bbx^*$ of \eqref{eq:empirical_min}.}
\label{fig_diagram}
\end{figure*}
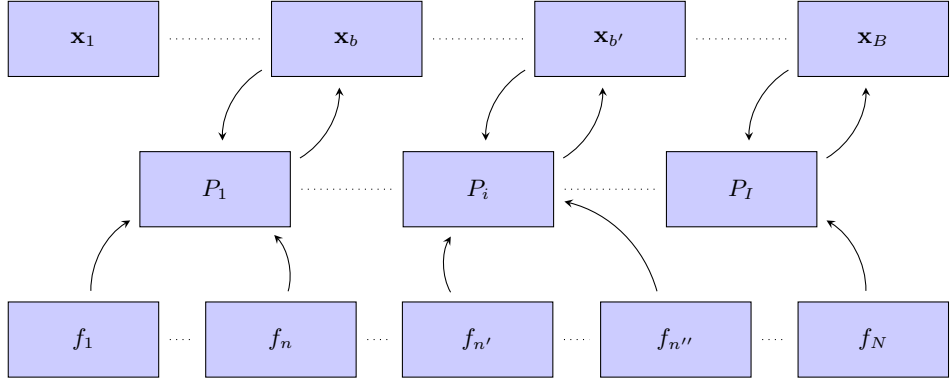

When $N$ and $p$ are large, the parallel processing architecture in Figure \ref{fig_diagram} becomes of interest. In this architecture, features are divided in $B$ blocks each of which contains $p_b\ll p$ features and a set of $I\ll B$ processors work in paralell on randomly chosen feature blocks while using a stocahstic subset of elements of the training set. In the schematic shown, Processor 1 fetches functions $f_1$ and $f_n$ to operate on block $\bbx_b$ and Processor $i$ fetches functions $f_{n'}$ and $f_{n''}$ to operate on block $\bbx_{b'}$. Other processors select other elements of the training set and other blocks with the majority of blocks remaining unchanged and the majority of functions remaining unused. The blocks chosen for update and the functions fetched for determination of block updates are selected independently at random in subsequent slots.

%

%
Problems that operate on blocks of the feature vectors {\it or} subsets of the training set, but not on both, blocks {\it and} subsets, exist. Block coordinate descent (BCD) is the generic name for methods in which the variable space is divided in blocks that are processed separately. Early versions operate by cyclically updating all coordinates at each step \cite{Tseng01convergenceof,xu2014globally}, while more recent parallelized versions of coordinate descent have been developed to accelerate convergence of BCD \cite{richtarik2012parallel,lu2013complexity,nesterov2012efficiency,wright2015coordinate,necoara2013parallel}. Closer to the architecture in Figure \ref{fig_diagram}, methods in which subsets of blocks are selected at random have also been proposed \cite{liu2013asynchronous}. BCD, serial, parallel, or random, can handle cases where the parameter dimension $p$ is large but requires access to all training samples at each iteration. 

Methods that utilize a subset of functions are known by the generic name of stochastic approximation and rely on the use of stochastic gradients. In plain stochastic gradient descent (SGD), the gradient of the aggregate function is estimated by the gradient of a randomly chosen function $f_n$ \cite{robbins1951}. Since convergence of SGD is slow more often that not, various recent developments have been aimed at accelerating convergence. These attempts include methodologies to reduce the variance of stochastic gradients \cite{schmidt2013minimizing,johnson2013accelerating,defazio2014saga} and the use of ideas from quasi-Newton optimization to handle difficult curvature profiles \cite{schraudolph2007stochastic,bordes2009sgd,mokhtari2014res,mokhtari2014global}. More pertinent to the work considered here are the use of cyclic block SGD updates \cite{xuyin2014} and the exploitation of sparsity properties of feature vectors to allow for parallel updates \cite{Niu11hogwild:a}. These methods are suitable when the number of elements in the training set $N$ is large but don't allow for parallel feature processing unless parallelism is inherent to the problem's structure. Moreover, parallel implementation of the SGD method can be considered when the dimension of the feature vectors is not massive and processors can update all the coordinates in a parallel manner \cite{yang2013parallel,daneshmand2015hybrid,facchinei2015parallel}.

The random parallel stochastic algorithm (RAPSA) proposed in this paper represents the first effort at implementing the architecture in Fig. \ref{fig_diagram} that randomizes over both, features and sample functions. In RAPSA, the functions fetched by a processor are used to compute the stochastic gradient component associated with a randomly chosen block (Section \ref{sec:rapsa}). The processors do not coordinate in either choice except to avoid selection of the same block. Our main technical contribution is to show that RAPSA iterates converge to the optimal classifier $\bbx^*$ when using a sequence of decreasing stepsizes and to a neighborhood of the optimal classifier when using constant stepsizes (Section \ref{sec:convergence_analysis}). In the latter case, we further show that the rate of convergence to this optimality neighborhood is linear in expectation. These results are interesting because only a subset of features are updated per iteration and the functions used to update different blocks are, in general, different. RAPSA is numerically evaluated on the MNIST digit recognition  problem (Section \ref{sec:simulations}).

%% file: diagram.tex
\def \thisplotscale {1}
\def \unit {\thisplotscale cm}

\tikzstyle{block} = [draw,
                     fill=blue!20,
                     minimum width  = 2*\unit,
                     minimum height = 1*\unit,]

{\small \begin{tikzpicture}[scale=\thisplotscale, shorten >=4,  shorten <=4]

    \node[block] at ( 1.75, 0) (processor 1) {$P_1$};
    \node[block] at ( 5.25, 0) (processor i) {$P_i$};
    \node[block] at ( 8.75, 0) (processor I) {$P_I$};            

    \node[block] at ( 0.0, 2) (variable 1)   {$\bbx_1$};
    \node[block] at ( 3.5, 2) (variable b)   {$\bbx_b$};
    \node[block] at ( 7.0, 2) (variable bp)  {$\bbx_{b'}$};
    \node[block] at (10.5, 2) (variable B)   {$\bbx_B$};            

    \node[block] at ( 0.000, -2) (function 1)   {$f_1$};
    \node[block] at ( 2.625, -2) (function n)   {$f_n$};
    \node[block] at ( 5.240, -2) (function np)   {$f_{n'}$};
    \node[block] at ( 7.875, -2) (function npp)   {$f_{n''}$};
    \node[block] at (10.500, -2) (function N)   {$f_N$};            

    \path (processor 1)   edge [dotted] (processor i);
    \path (processor i)   edge [dotted] (processor I);

    \path (variable 1)   edge [dotted] (variable b);
    \path (variable b)   edge [dotted] (variable bp);
    \path (variable bp)  edge [dotted] (variable B);
    
    \path (function 1)   edge [dotted] (function n);
    \path (function n)   edge [dotted] (function np);
    \path (function np)  edge [dotted] (function npp);
    \path (function npp) edge [dotted] (function N);

    \path[-stealth] (variable b)   edge [bend right] (processor 1);
    \path[-stealth] (processor 1)  edge [bend right] (variable b);
    \path[-stealth] (function 1)   edge [bend left]  (processor 1);
    \path[-stealth] (function n)   edge [bend right] (processor 1);    

    \path[-stealth] (variable bp)  edge [bend right] (processor i);
    \path[-stealth] (processor i)  edge [bend right] (variable bp);
    \path[-stealth] (function np)  edge [bend left]  (processor i);
    \path[-stealth] (function npp) edge [bend right] (processor i);    

    \path[-stealth] (variable B)   edge [bend right] (processor I);
    \path[-stealth] (processor I)  edge [bend right] (variable B);
    \path[-stealth] (function N)   edge [bend right]  (processor I);

\end{tikzpicture}} 

%% file: Problem.tex
%

\section{Random Parallel Stochastic Algorithm (RAPSA)}\label{sec:rapsa}

%

We consider a more general formulation of \eqref{eq:empirical_min} in which the number $N$ of functions $f_n$ is not necessarily finite. Introduce then a random variable $\bbtheta \in \bbTheta \subset \reals^q$ that determine the choice of the random smooth convex function $f(\cdot,\bbtheta) : \reals^p \to \reals$. We consider the problem of minimizing the expectation of the random functions $F(\bbx):=\mbE_{\bbtheta}[f(\bbx,\bbtheta)]$, 
\begin{equation} \label{eq:block_stoch_opt}
   \bbx^* := \argmin_{\bbx} F(\bbx)
          := \argmin_{\bbx} \mbE_{\bbtheta}\left[{f(\bbx , \bbtheta)}\right].
\end{equation}
Problem \eqref{eq:empirical_min} is a particular case of \eqref{eq:block_stoch_opt} in which each of the functions $f_n$ is drawn with probability $1/N$. We refer to $f(\cdot,\bbtheta)$ as instantaneous functions and to $F(\bbx)$ as the average function. 

RAPSA utilizes $I$ processors to update a random subset of blocks of the variable $\bbx$, with each of the blocks relying on a subset of randomly and independently chosen elements of the training set; see Figure \ref{fig_diagram}. Formally, decompose the variable $\bbx$ into $B$ blocks to write $\bbx=[\bbx_1;\dots;\bbx_B]$, where block $b$ has length $p_b$ so that we have $\bbx_b \in \reals^{p_b}$. At iteration $t$, processor $i$ selects a random index $b_i^t$ for updating and a random subset $\bbTheta_i^t$ of $L$ instantaneous functions. It then uses these instantaneous functions to determine stochastic gradient components for the subset of variables $\bbx_b=\bbx_{b_i^t}$ as an average of the components of the gradients of the functions $f( \bbx^{t}, \bbtheta)$ for $\bbtheta\in\bbTheta_i^t$,
\begin{equation}\label{block_sto_grad}
    \nabla_{\bbx_b} f( \bbx^{t}, \bbTheta_i^t)
      = \frac{1}{L}\sum_{\bbtheta\in \bbTheta_i^t} \nabla_{\bbx_b} f( \bbx^{t}, \bbtheta), 
    \qquad   b = b_i^t.
\end{equation}
The stochastic gradient block in \eqref{block_sto_grad} is then modulated by a possibly time varying stepsize $\gamma^t$ and used by processor $i$ to update the block $\bbx_b=\bbx_{b_i^t}$ 
\begin{align} \label{eq:block_stochastic_gradient_1}
   \bbx^{t+1}_{b}  =  \bbx^{t}_b  - \gamma^t  \nabla_{\bbx_b} f( \bbx^{t}, \bbTheta_i^t) 
   \qquad   \qquad   b = b_i^t .
\end{align}
RAPSA is defined by the joint implementation of \eqref{block_sto_grad} and \eqref{eq:block_stochastic_gradient_1} across all $I$ processors. The selection of blocks is coordinated so that no processors operate in the same block. The selection of elements of the training set is uncoordinated across processors. The fact that at any point in time a random subset of blocks is being updated utilizing a random subset of elements of the training set means that RAPSA requires almost no coordination between processors. The contribution of this paper is to show that this very lean algorithm converges to the optimal argument $\bbx^*$ as we show in the following section.

%% file: Convergence.tex

%
\section{Convergence Analysis}\label{sec:convergence_analysis}

We show in this section that the sequence of objective function values $F(\bbx^t)$ generated by RAPSA approaches the optimal objective function value $F(\bbx^{*})$. In establishing this result we define the set $\ccalS^t$ containing the blocks that are updated at step $t$ with associated indices $\ccalI^t \subset \{1,\dots,B\}$. Note that components of the set $\ccalS^t$ are chosen uniformly at random from the set of blocks $\{\bbx_1,\dots,\bbx_B\}$. The definition of $\ccalS^t$ is such that the time evolution of RAPSA iterates can be written as, [cf. \eqref{eq:block_stochastic_gradient_1}],
\begin{align} \label{eq:block_stochastic_gradient}
   {\bbx^{t+1}_{i}  \ = \ \bbx^{t}_{i}  - \gamma^t\  \nabla_{\bbx_i} f( \bbx^{t}, \bbTheta_i^t) \qquad \forall \ \bbx_i\in \ccalS^t,} 
\end{align}
while the rest of the blocks remain unchanged, i.e., $ \bbx^{t+1}_{i}=\bbx^{t}_{i}$ for $\bbx_i\notin \ccalS^t$. Since the number of updated blocks is equal to the number of processors, the ratio of updated blocks is $r:=|\ccalI^t|/B=I/B$. 

To prove convergence of RAPSA, we require the following assumptions

%
\begin{assumption}\label{convexity_assumption} 
The instantaneous objective functions $f(\bbx,\bbtheta)$ are differentiable and the average function $F(\bbx)$ is strongly convex with parameter $m>0$.
%
\end{assumption}

%
\begin{assumption}\label{Lipschitz_assumption} The average objective function gradients $\nabla F(\bbx)$ are Lipschitz continuous with respect to the Euclidian norm with parameter $M$. I.e., for all $\bbx, \hbx \in \reals^p$, it holds
\begin{equation}
   \| \nabla F(\bbx)-\nabla F(\hbx) \| \ \leq\  M\ \| \bbx- \hbx \|.
\end{equation}
\end{assumption}

%
\begin{assumption}\normalfont\label{ass_bounded_stochastic_gradient_norm} The second moment of the norm of the stochastic gradient is bounded for all $\bbx$, i.e., there exists a constant $K$ such that for all variables $\bbx$, it holds
\begin{equation}\label{ekhtelaf}
   \mbE_{\bbtheta} \big{[} \|\nabla f( \bbx^t, \bbtheta^t)\|^{2} \given{\bbx^t}\big{]} \leq K.
\end{equation} \end{assumption}


Notice that Assumption \ref{convexity_assumption} only enforces strong convexity of the average function $F$, while the instantaneous functions $f_i$ may not be even convex. Further, notice that since the instantaneous functions $f_i$ are differentiable the average function $F$ is also differentiable. The Lipschitz continuity of the average function gradients $\nabla F$ is customary in proving objective function convergence for descent algorithms. The restriction imposed by Assumption \ref{ass_bounded_stochastic_gradient_norm} is a standard condition in stochastic approximation literature \cite{robbins1951}, its intent being to limit the variance of the stochastic gradients \cite{Nemirovski}. 

Our first result comes in the form of a expected descent lemma that relates the expected difference of subsequent iterates to the gradient of the average function.

\begin{lemma}\label{exp_wrt_blocks}
Consider the random parallel stochastic algorithm defined in \eqref{block_sto_grad}-\eqref{eq:block_stochastic_gradient}. Recall the definitions of the set of updated blocks $\ccalI^t$ which are randomly chosen from the total $B$ blocks. 
Define $\ccalF^t$ as a sigma algebra that measures the history of the system up until time $t$. Then, the expected value of the difference $\bbx^{t+1}-\bbx^{t}$ with respect to the random set $\ccalI^t$ given $\ccalF^t$ is 
\begin{equation}\label{lemma_RAPS_dec_claim_1}
\mathbb{E}_{\ccalI^t}\!\left[{\bbx^{t+1}-\bbx^{t}\mid \ccalF^t}\right] =
	 - r\gamma^t \ \nabla f( \bbx^{t}, \bbTheta^t).
\end{equation}
Moreover, the expected value of the squared norm $\|\bbx^{t+1}-\bbx^{t}\|^2$ with respect to the random set $\ccalS^t$ given $\ccalF^t$ can be simplified as
\begin{equation}\label{lemma_RAPS_dec_claim_2}
\mathbb{E}_{\ccalI^t}\!\left[\|{\bbx^{t+1}-\bbx^{t}\|^2\mid \ccalF^t}\right] = 
	 {r(\gamma^t)^2 }\ \left\|\nabla f( \bbx^{t}, \bbTheta^t)\right\|^2.
\end{equation}
\end{lemma}

\begin{myproof}
See Appendix A.
\end{myproof}

Notice that in the regular stochastic gradient descent method the difference of two consecutive iterates $\bbx^{t+1}-\bbx^{t}$ is equal to the stochastic gradient $\nabla f( \bbx^{t}, \bbTheta^t)$ times the stepsize $\gamma^t$. Based on the first result in Lemma \ref{exp_wrt_blocks}, the expected value of stochastic gradients with respect to the random set of blocks $\ccalI^t$ is the same as the one for SGD except that it is multiplied by the fraction of updated blocks $r$. Expression in \eqref{lemma_RAPS_dec_claim_2} shows the same relation for the expected value of the squared difference $\|\bbx^{t+1}-\bbx^{t}\|^2$. These relationships confirm that in expectation RAPSA behaves as SGD which allows us to establish the global convergence of RAPSA.

\begin{proposition}\label{martingale_prop}
Consider the random parallel stochastic algorithm defined in \eqref{block_sto_grad}-\eqref{eq:block_stochastic_gradient}. 
If Assumptions \ref{convexity_assumption}-\ref{ass_bounded_stochastic_gradient_norm} hold, then the objective function error sequence $F(\bbx^t)-F(\bbx^*)$ satisfies 
\begin{align}\label{martingale_prop_claim}
&\mathbb{E}\left[F(\bbx^{t+1})-F(\bbx^*)\mid \ccalF^t\right]\nonumber\\
&\quad	\leq 
	\left( 1- {2m r\gamma^t}{} \right)\left( F(\bbx^{t}) -F(\bbx^*)\right)+ \frac{rM  K(\gamma^t)^2}{2 }.
\end{align}
\end{proposition}

\begin{myproof}
See Appendix B.
\end{myproof}


Proposition \ref{martingale_prop} leads to a supermartingale relationship for the sequence of objective function errors $F(\bbx^t)-F(\bbx^*)$. In the following theorem we show that if the sequence of stepsize satisfies standard stochastic approximation diminishing step-size rules (non-summable and squared summable), the sequence of objective function errors $F(\bbx^t)-F(\bbx^*)$ converges to null almost surely. Considering the strong convexity assumption this result implies almost sure convergence of the sequence $\| \bbx^{t}-\bbx^{*} \|^{2}$ to null.

\begin{theorem}\label{RAPSA_convg_thm}
Consider the random parallel stochastic algorithm defined in \eqref{block_sto_grad}-\eqref{eq:block_stochastic_gradient}. If Assumptions \ref{convexity_assumption}-\ref{ass_bounded_stochastic_gradient_norm} hold true and the sequence of stepsizes are non-summable $\sum_{t=0}^\infty \gamma^t=\infty$ and square summable $\sum_{t=0}^\infty (\gamma^t)^2<\infty$, then sequence of the variables $\bbx^t$ generated by RAPSA converges almost surely to the optimal argument $\bbx^*$, 
\begin{equation}\label{rapsa_as_convg}
\lim_{t\to \infty} \|\bbx^t-\bbx^*\|^2 \ =\ 0 \qquad a.s.
\end{equation}
Moreover, if stepsize is defined as $\gamma^t:=\gamma^0T^0/(t+T^0)$ and the stepsize parameters are chosen such that $2mr \gamma^0 T^0>1$, then the expected average function error $\E{F(\bbx^t)-F(\bbx^*)}$ converges to null at least with a sublinear convergence rate of order $O(1/t)$,
\begin{equation}\label{rapsa_rate}
\E{F(\bbx^t)-F(\bbx^*)} \ \leq \ \frac{ C}{t+T^0},
\end{equation}
where the constant $C$ is defined as 
\begin{equation}\label{Constant_C}
C= \max\left\{\frac{rMK (\gamma^0 T^0)^2}{4mr\gamma^0 T^0-2},\ T^0(F(\bbx^0)-F(\bbx^*))\right\}.
\end{equation}
\end{theorem}
\begin{myproof}
See Appendix C.
\end{myproof}

The result in Theorem \ref{RAPSA_convg_thm} shows that when the sequence of stepsize is diminishing as $\gamma^t=\gamma^0T^0/(t+T^0)$, the average objective function value $F(\bbx^t)$ sequence converges to the optimal objective value $F(\bbx^*)$ with probability 1.\footnote{The expectation on the left hand side of \eqref{rapsa_rate}, and throughout the subsequent convergence rate analysis, is taken with respect to the algorithm history $\ccalF_0$, which contains all randomness in both $\bbTheta_t$ and $\ccalI_t$ for all $t\geq 0$.} Further, the rate of convergence in expectation is at least in the order of $O(1/t)$. 
Diminishing stepsizes are useful when exact convergence is required, however, for the case that we are interested in a specific accuracy $\eps$ the more efficient choice is using a constant stepsize. In the following theorem we study the convergence properties of RAPSA for a constant stepsize $\gamma^t=\gamma$.

\begin{theorem}\label{RAPSA_convg_thm_finite}
Consider the random parallel stochastic algorithm defined in \eqref{block_sto_grad} and \eqref{eq:block_stochastic_gradient}. If Assumptions \ref{convexity_assumption}-\ref{ass_bounded_stochastic_gradient_norm} hold true and the stepsize is constant $\gamma^t=\gamma$, then a subsequence of the variables $\bbx^t$ generated by RAPSA converges almost surely to a neighborhood of the optimal argument $\bbx^*$ as
\begin{equation}\label{rapsa_as_convg_finite}
\liminf_{t\to \infty}\ F(\bbx^t)-F(\bbx^*)\ \leq\ \frac{\gamma M K}{4m} \qquad a.s.
\end{equation}
Moreover, if the constant stepsize $\gamma$ is chosen such that $2m r\gamma<1$ then the expected average function value error $\E{F(\bbx^t)-F(\bbx^*)}$ converges \textit{linearly} to an error bound as
\begin{align}\label{rapsa_rate_finite}
\E{F(\bbx^{t})-F(\bbx^*)}
	&\leq 
	\left( 1- {2m\gamma r}{} \right)^{t}( F(\bbx^{0}) -F(\bbx^*))
	\nonumber\\
	&\quad
	+ \frac{\gamma M  K}{4 m}.
\end{align}
\end{theorem}
\begin{myproof}
See Appendix D.
\end{myproof}

\begin{figure*}
\centering
\setcounter{subfigure}{0}
\begin{subfigure}{0.8\columnwidth}
\includegraphics[width=1\linewidth, height = 0.65\linewidth]
                	{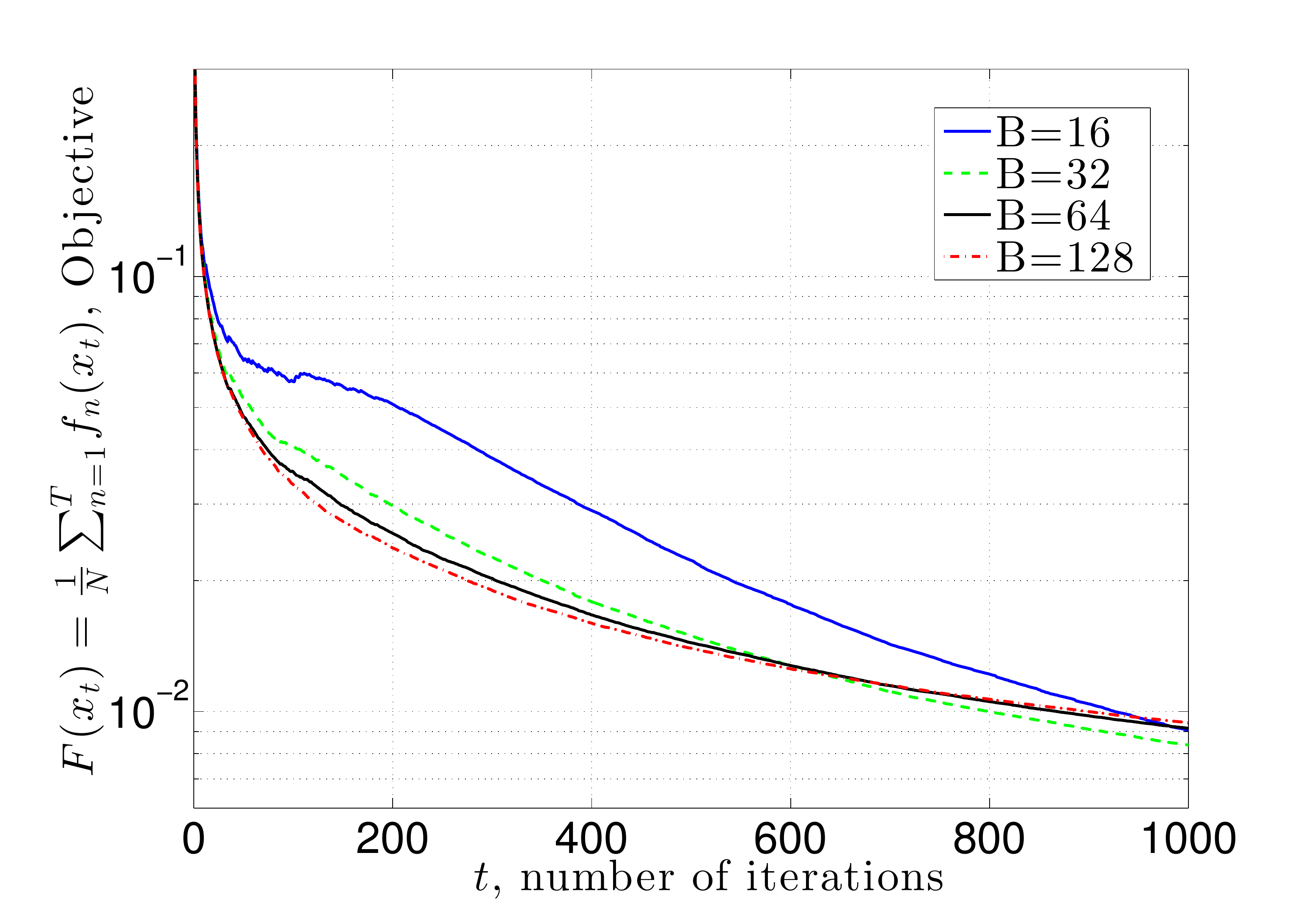}
\caption{Objective $F(\bbx^t)$ vs. iteration $t$}
\label{subfig:linear_a}
\end{subfigure} \hspace{15mm}
\begin{subfigure}{0.8\columnwidth}
\includegraphics[width=1\linewidth, height = 0.65\linewidth]
                {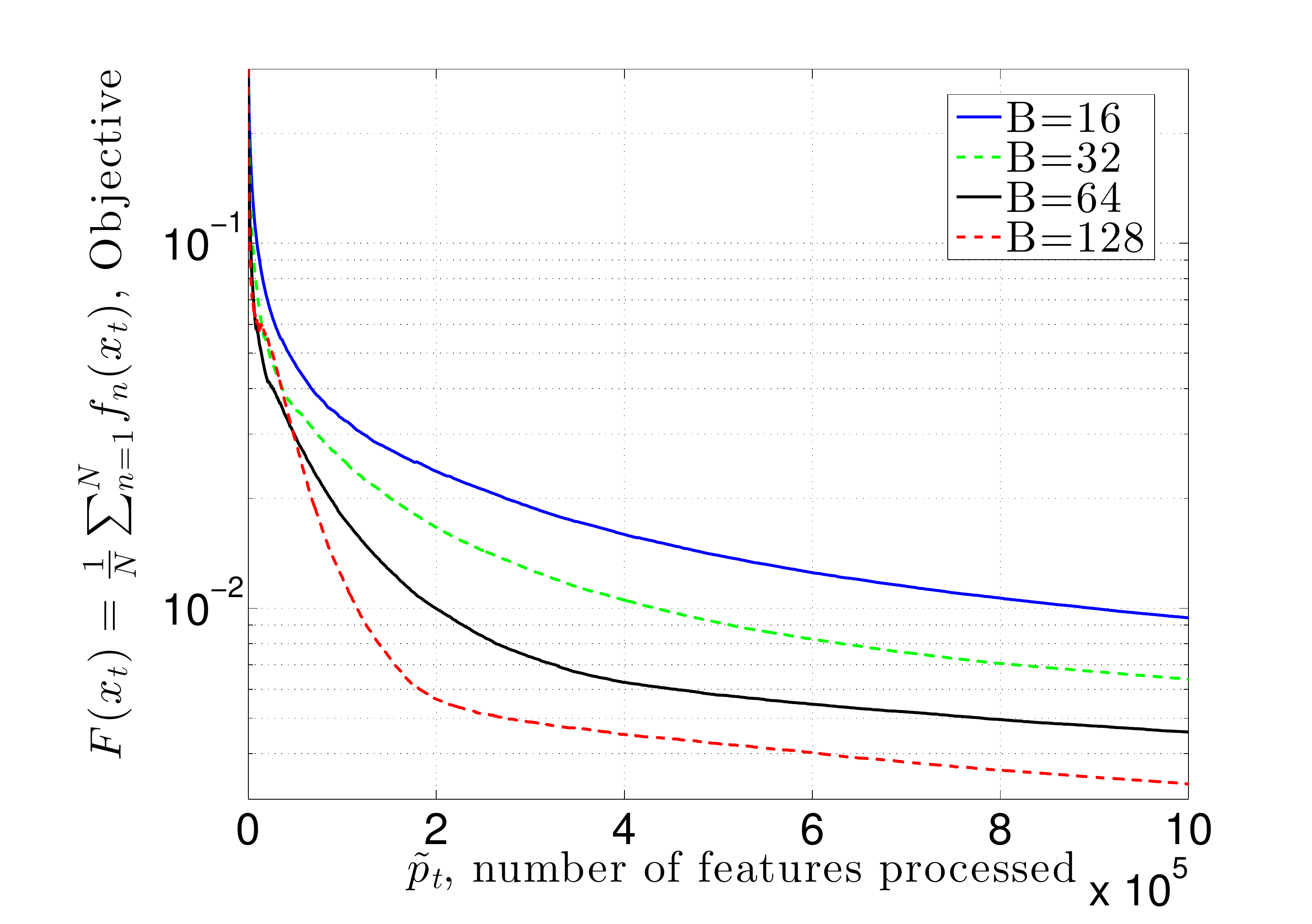}
\caption{Objective $F(\bbx^t)$ vs. feature $\tilde{p}_t$ }
\label{subfig:linear_b}
\end{subfigure}\vspace{-2mm}
\caption{RAPSA on a linear regression (quadratic minimization) problem with signal dimension $p=1024$ for $N=10^4$ iterations without mini-batching $L=1$ for different number of blocks $B=\{16,32,64,128\}$. We use hybrid step-size $\gamma^t= \min(10^{-3},10^{-3} \tilde{T}_0/t)$ with annealing rate $\tilde{T}_0=450$. Convergence is comparable across the different cases in terms of number of iterations, but in terms of number of features processed $B=128$ has the best performance and $B=16$ has the worst performance. This shows that updating less features/coordinates per iterations leads to faster convergence in terms of number of processed features.} \label{fig:rapsa_linear_regression}
\end{figure*}
Notice that according to the result in \eqref{rapsa_rate_finite} there exits a trade-off between accuracy and speed of convergence. Decreasing the constant stepsize $\gamma$ leads to a smaller error bound ${\gamma M  K}/{4 m}$ and a more accurate convergence, while the linear convergence constant $\left( 1- {2m\gamma r}{} \right)$ increases and the convergence rate becomes slower. Further, note that the error of convergence ${\gamma M  K}/{4 m}$ is independent of the ratio of updated blocks $r$, while the constant of linear convergence $1- {2m\gamma r}$ depends on $r$. Therefore, updating a fraction of the blocks at each iteration decreases the speed of convergence for RAPSA relative to SGD that updates all of the blocks, however, both of the algorithms reach the same accuracy. 

To achieve accuracy $\eps$ the sum of two terms in the right hand side of \eqref{rapsa_rate_finite} should be smaller than $\eps$. Let's consider $\phi$ as a positive constant that is strictly smaller than $1$, i.e., $0<\phi<1$. Then, we want to have
\begin{equation}\label{condition_on_gamma}
 \frac{\gamma M  K}{4 m}\leq \phi \eps , \quad\!\! \left( 1- {2m\gamma r}{} \right)^{t}( F(\bbx^{0}) -F(\bbx^*)) \leq (1-\phi)\eps.
\end{equation}
Therefore, to satisfy the first condition in \eqref{condition_on_gamma} we set the stepsize as $\gamma=4m\phi\eps/MK$. Apply this substitution into the second inequality in  \eqref{condition_on_gamma} and consider the inequality $a+\ln(1-a)<0$  for $0<a<1$, to obtain that 
\begin{equation}\label{num_ite}
t\geq\frac{MK}{8m^2r\phi \eps}\ln\left(\frac{ F(\bbx^{0}) -F(\bbx^*)}{(1-\phi)\eps}\right).
\end{equation}
The lower bound in \eqref{num_ite} shows the minimum number of required iterations for RAPSA to achieve accuracy $\eps$.

%% file: Simulations.tex
\section{Numerical analysis}
\label{sec:simulations}

In this section we study the numerical performance of the doubly stochastic approximation algorithm developed in Section \ref{sec:rapsa} by first considering a linear regression problem. We then use RAPSA to develop an automated decision system to distinguish between distinct hand-written digits. 

\subsection{Linear Regression}\label{subsec:lmmse}

We consider a setting in which observations $\bbz_n\in\reals^q$ are collected which are assumed to be noisy linear transformations $\bbz_{n}=\bbH_n \bbx + \bbw_{n}$ of a signal $\bbx\in\reals^p$ which we would like to estimate, and $\bbw \sim \ccalN(0, \sigma^2 I_q)$ is a Gaussian random variable. For a finite set of samples $N$, the optimal $\bbx^*$ is computed as the least squares estimate $\bbx^*:=\argmin_{\bbx\in\reals^p } ({1}/{N})\sum_{n=1}^N \|\bbH_{n}\bbx- \bbz_{n}\|^2$. We run RAPSA on LMMSE estimation problem instances where $q=1$, $p=1024$, and $N=10^4$ samples are given. The observation matrices $\bbH_n \in \reals^{q\times p}$ are chosen as $p$-dimensional Gaussian vectors, the true signal $\bbx=(1/4) \bbone$, and the noise variance $\sigma^2=10^{-1.5}$. We assume that the number of processors $I=16$ is fixed and each processor is in charge of $1$ block. We consider different number of blocks $B=\{16,32,64,128\}$. Note that when the number of blocks is $B$, there are $p/B=1024/B$ coordinates in each block. 

\begin{figure*}
\begin{subfigure}{.67\columnwidth}
\includegraphics[width=\linewidth, height = 0.7\linewidth]
		{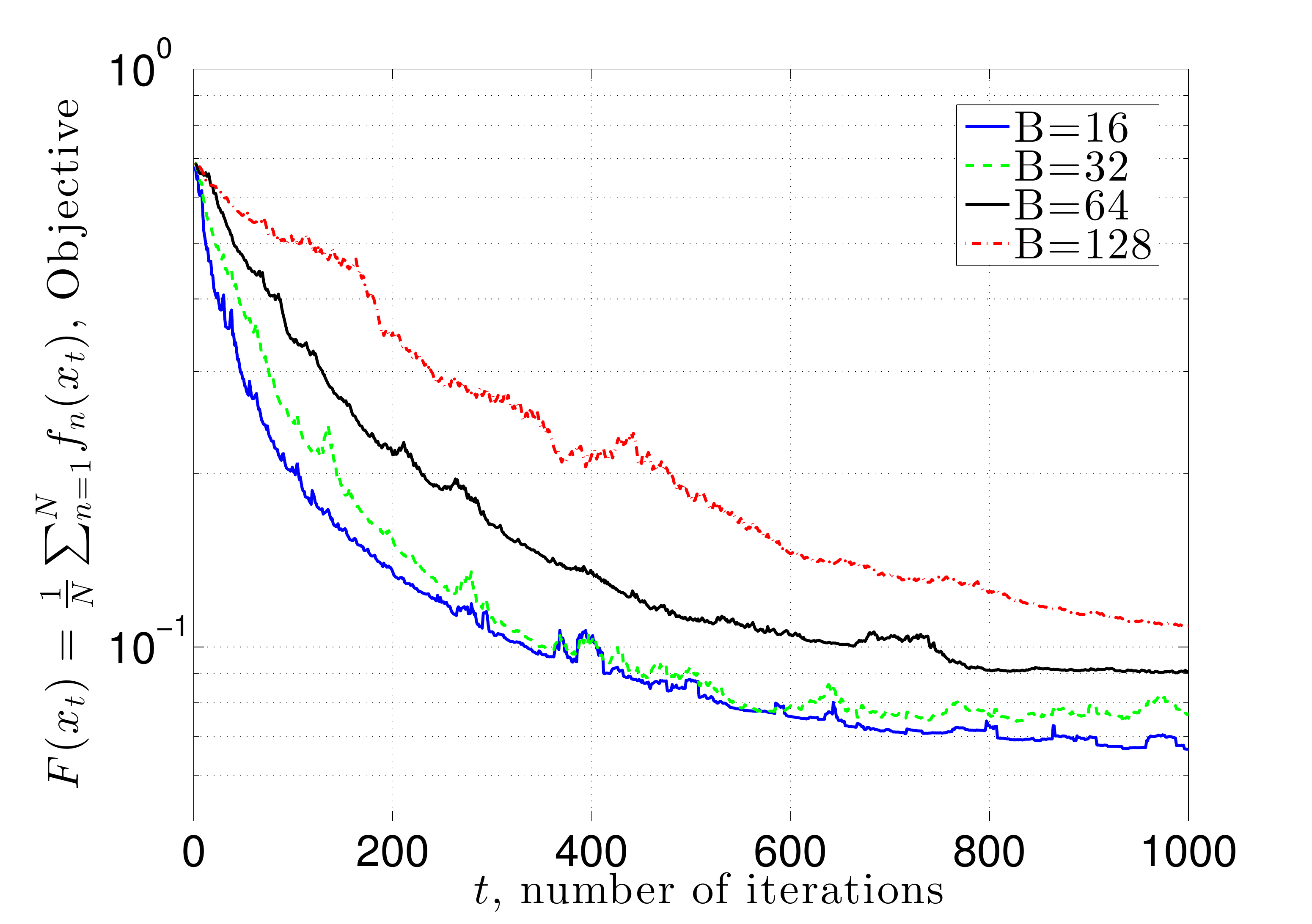}
\caption{Objective $F(\bbx^t)$ vs. iteration $t$.}
\label{subfig:mnist1}
\end{subfigure}
\begin{subfigure}{.67\columnwidth}
\includegraphics[width=\linewidth, height = 0.7\linewidth]
                {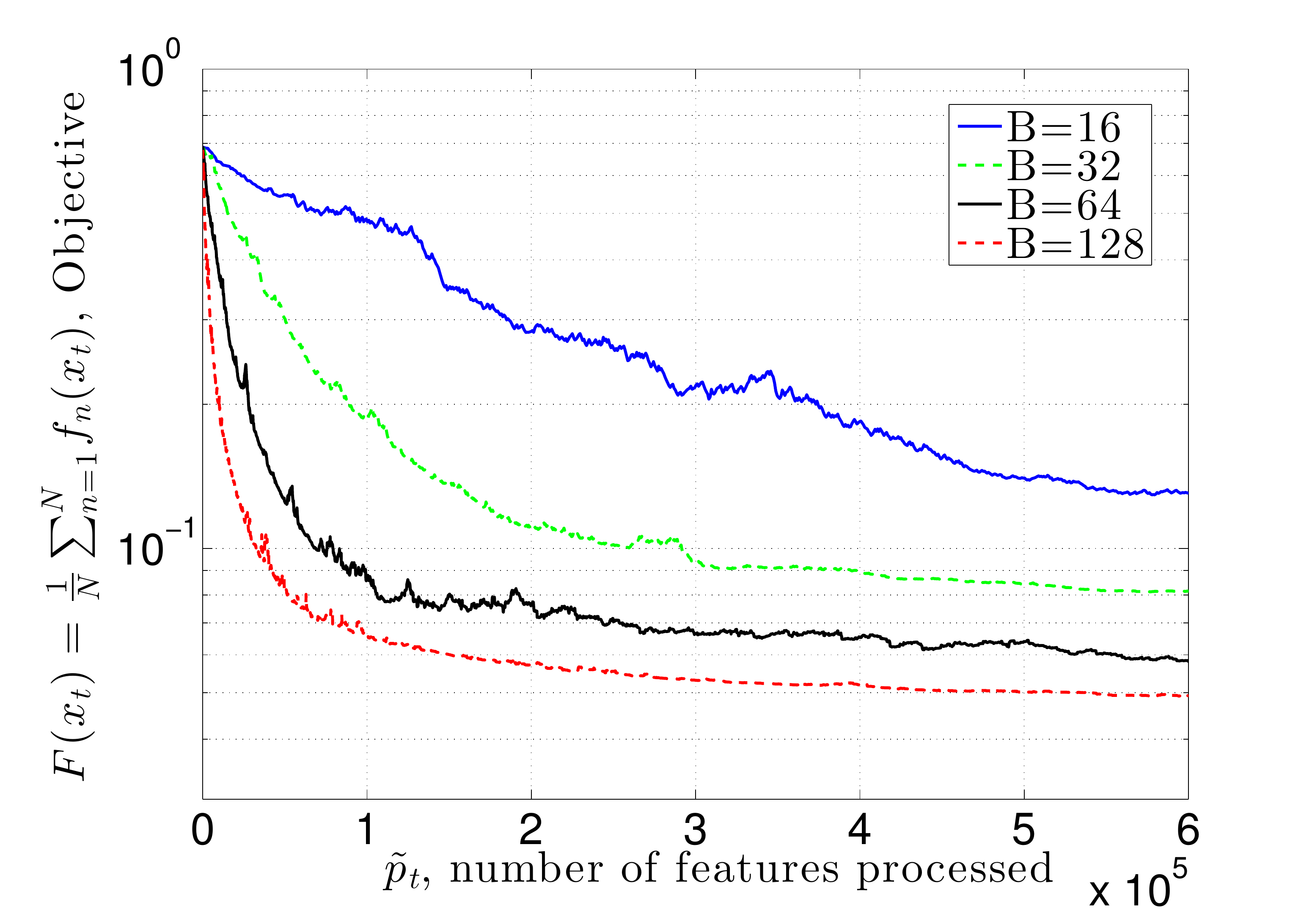}
\caption{Objective $F(\bbx^t)$ vs. feature $\tilde{p}_t$.}
\label{subfig:mnist2}
\end{subfigure} 
\begin{subfigure}{.67\columnwidth}
\includegraphics[width=\linewidth,height = 0.7\linewidth]
                {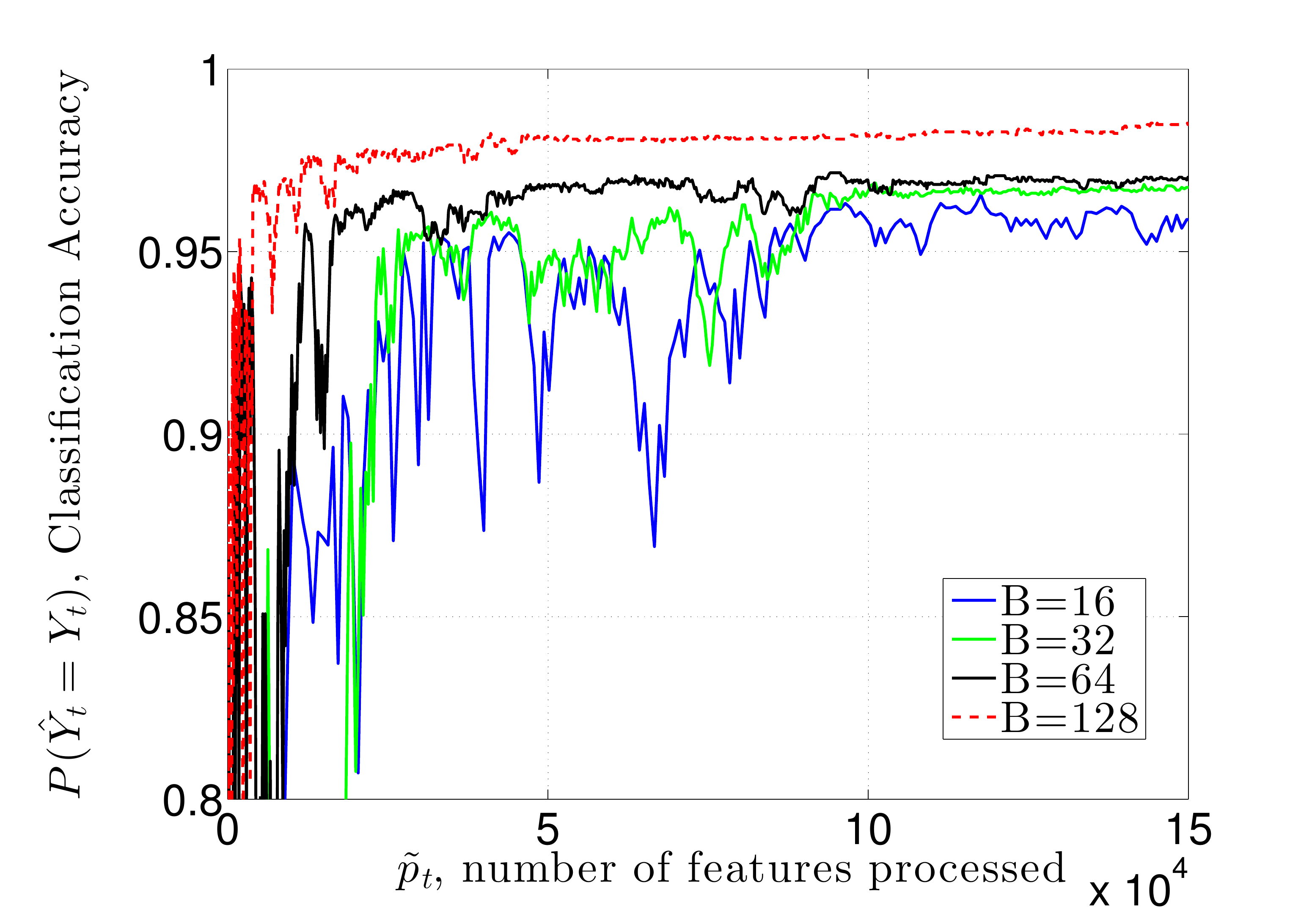}
\caption{Test set accuracy vs. feature $\tilde{p}_t$.}
\label{subfig:mnist3}
\end{subfigure}\vspace{-1mm}
\caption{RAPSA on MNIST data. Algorithm performance is comparable across different numbers of decision variable coordinates updated per iteration $t$, but in terms of number of features processed, RAPSA performance best when using the least features per update.} \label{fig:rapsa_mnist}
\end{figure*}

We use of a hybrid algorithm step-size  $\gamma^t= \min(\eps, \eps\tilde{T}_0/t)$ which is a constant $\eps=10^{-3}$ for the first $\tilde{T}_0=500$ iterations, after which it diminishes as $O(1/t)$. The size of mini-batch is set as $L=1$ in the subsequent experiment. To determine the advantages of incomplete randomized parallel processing, we vary the number of coordinates updated at each iteration. In the case that $B=16$, $B=32$, $B=64$, and $B=128$ the number of updated features per iterations are $1024$, $512$, $256$, and $128$, respectively. Notice that the case that $B=16$ can be interpreted as parallel SGD since all the coordinates are updated per iteration, while in other cases $B>16$ only a subset of $1024$ coordinates are updated. 

 Fig. \ref{subfig:linear_a} illustrates the convergence path of RAPSA's objective value $F(\bbx^t)=({1}/{N})\sum_{n=1}^N \|\bbH_{n}\bbx^t- \bbz_{n}\|^2$ versus the number of iterations $t$. We observe that the algorithm performance is comparable across different number of updating coordinates per iteration. However, comparing algorithm performance over iteration $t$ across varying numbers of blocks updates is unfair. If RAPSA is run on a problem for which $B=32$, then at iteration $t$ it has only processed {\it half} the data that parallel SGD, i.e., $B=16$, has processed by the same iteration. Thus for completeness we also consider the algorithm performance in terms of number of features processed $\tdp_t$ which is given by $\tdp_t=ptI/B$.

In Fig. \ref{subfig:linear_b}, we display the convergence of the mean square error $F(\bbx^t)$ in terms of number of features processed $\tilde{p}_t$. In doing so, we may clearly observe the advantages of updating fewer features/coordinates per iteration. That is, to achieve the benchmark $F(\bbx^t)\leq 10^{-2}$, we require $\tdp_t=8.98\times 10^5$, $\tdp_t=4.33\times 10^5$, $\tdp_t=1.99\times 10^5$, and $\tdp_t=1.15\times 10^5$ processed features, respectively, for the cases that the number of blocks are $B=16$, $B=32$, $B=64$, and $B=128$. This result shows that updating fewer coordinates per iteration yields substantial convergence gains in terms of number of features processed. This advantage comes from the advantage of Gauss-Seidel method with respect to Jacobi algorithm.

Consider one path over the dataset for the cases that $B=16$ and $B=32$. In $B=16$, given $\bbx^t$, we update all the $1024$ coordinates/features to update $\bbx^t$ and compute $\bbx^{t+1}$. On the other hand, for the case that $B=32$, we update $512$ features of $\bbx^t$ and get $\bbx^{t+1}$. Then we use the updated variable $\bbx^{t+1}$ to update the rest of coordinates and compute $\bbx^{t+2}$. This observation shows that $B=16$ acts as a Jacobi method, updating all coordinates of $\bbx$ in parallel, while $B>16$ has the structure of Gauss-Seidel method makes use of the updated information as it proceeds to update different coordinates. The superior behavior of Gauss-Seidel techniques as compared to Jacobi methods is well known, and underlies the performance gains in Fig. \ref{subfig:linear_b}.

%
\subsection{Hand-Written Digit Recognition}\label{subsec:mnist}
We now make use of RAPSA for digit classification. To do so, let $\bbz\in\reals^p$ be a feature vector encoding pixel intensities of an image, and let $y\in \{-1,1\}$ be an indicator variable of whether the image contains the digit $0$ or $8$, in which case the binary indicator is respectively $y=-1$ or $y=1$.
We model the task of learning a hand-written digit detector as a logistic regression problem, where one aims to train a classifier $\bbx \in \reals^p$ to determine the relationship between feature vectors $\bbz_n \in \reals^p$ and their associated labels $y_n \in \{-1,1\}$  for $n=1,\dots,N$. 
%
The empirical risk minimization associated with training set $\ccalT=\{(\bbz_n,y_n)\}_{n=1}^N$ is to find $\bbx^*$ as the maximum likelihood estimate
\begin{equation}\label{logistic_regression}
\bbx^* := \argmin_{\bbx\in\reals^p}  \frac{\lambda}{2}\|\bbx\|^2 + \frac{1}{N}\sum_{n=1}^N \log (1 + \exp({-y_n \bbx^T \bbz_n}))\; ,
\end{equation}
where the regularization term $({\lambda}{/2})\|\bbx\|^2$ is added to avoid overfitting. We use the MNIST dataset \cite{citeulike:599493}, in which feature vectors $\bbz_n \in \reals^p$ are $p=28^2=784$ pixel images whose values are recorded as intensities, or elements of the unit interval $[0,1]$. Considered here is the subset associated with digits $0$ and $8$, a training set $\ccalT=\{\bbz_n,y_n\}_{n=1}^N$ with $N=1.76\times 10^4$ sample points. 

We run RAPSA on this training subset for the cases that $B=16$, $B=32$, $B=64$, and $B=128$, which are associated with updating $p$, $p/2$, $p/4$, and $p/8$ features per iteration. In Fig. \ref{fig:rapsa_mnist} we show the result of running RAPSA for this logistic regression problem with hybrid step-size $\gamma^t= \min(10^{-2.5},10^{-2.5} \tilde{T}_0/t)$, with $\tilde{T}_0=525$ and no mini-batching $L=1$. We observe in Fig. \ref{subfig:mnist1} that using full stochastic gradients is better than only updating \emph{some} of the coordinates in terms of the number of iterations $t$. In particular, to reach the objective benchmark $F(\bbx^t) \leq 10^{-1}$, we have to run RAPSA $t=335$, $t=354$, and $t=741$ iterations, for the cases that $B=16$, $B=32$, and $B=64$, respectively. For $B=128$ the objective function $F(\bbx^t) \leq 10^{-1}$ is not reached after $1000$ iterations. 

However, as previously noted, iteration index $t$ is an unfair comparator for objective convergence since the four different setting process different number of features per iteration. Hence, we illustrate  the objective $F(\bbx^t)$ vs. feature $\tilde{p}_t$ in Fig. \ref{subfig:mnist2}. Here we recover the advantages of randomized incomplete parallel processing: updating fewer blocks per iteration yields improved algorithm performance. We observe using the least amount of information per iteration yields the fastest convergence in terms of number of features processed. 

We also consider the algorithm's classification accuracy on a test subset of size $\tilde{N}=5.88\times10^3$, the results of which are shown in Fig. \ref{subfig:mnist3}. In terms of number of features processed $\tilde{p}_t$, we see that the result for classification accuracy on a test set is consistent with the results for the convergence of the objective function value. 
 

%% file: Conclusions.tex

\section{Conclusions}\label{sec_conclusions}

The random parallel stochastic algorithm (RAPSA) pro- posed in this paper as a doubly stochastic algorithm. RAPSA is doubly stochastic since each processors utilizes a random set of functions to compute the stochastic gradient associated with a randomly chosen sets of variable coordinates. We showed the proposed algorithm converges to the optimal solution sublinearly when the stepsize is diminishing. Moreover, linear convergence to a neighborhood of the optimal solution can be achieved using a constant stepsize. A detailed comparison between RAPSA and parallel SGD for training a quadratic program and a logistic regressor is provided. The numerical results showcase the advantage of RAPSA with respect to parallel SGD.

%% file: Appendix.tex

\section*{APPENDIX}

\subsection{Proof of Lemma \ref{exp_wrt_blocks}}\label{apx_lemma_exp_wrt_blocks}

Recall that the components of vector $\bbx^{t+1}$ are equal to the components of $\bbx^t$ for the coordinates that are not updated at step $t$, i.e., $i\notin \ccalI^t$. For the updated coordinates $i\in \ccalI^t$ we know that $\bbx^{t+1}_{i}=\bbx^{t}_{i}-\gamma^t  \nabla_{\bbx_i^t} f( \bbx^{t}, \bbtheta^t)$. Therefore, $B-I$ blocks of the vector $\bbx^{t+1}-\bbx^t$ are 0 and the remaining $I$ randomly chosen blocks are given by $-\gamma^t  \nabla_{\bbx_i^t} f( \bbx^{t}, \bbtheta^t)$. Notice that there are ${B}\choose{I}$ different ways for picking $I$ blocks out of the whole $B$ blocks. Therefore, the probability of each combination of blocks is $1/ {{B}\choose{I}}$. Further, each block appears in ${B-1}\choose{I-1}$ of the combinations. Therefore, the expected value can be written as 
\begin{equation}\label{lemma_RAPS_dec_20}
\mathbb{E}_{\ccalI^t}\!\left[   \bbx^{t+1}-\bbx^t \mid \ccalF^t   \right]
	=\frac{{{B-1}\choose{I-1}}}{{{m}\choose{I}}} \left(  -\gamma^t  \nabla f( \bbx^{t}, \bbTheta^t)  \right).
\end{equation}
Observe that simplifying the ratio in the right hand sides of \eqref{lemma_RAPS_dec_20} leads to 
\begin{equation}\label{lemma_RAPS_dec_30}
\frac{{{B-1}\choose{I-1}}}{{{B}\choose{I}}} =\frac{\frac{(B-1)!}{(I-1)!\times (B-I)!}}{\frac{p!}{I!\times (B-I)!}}=\frac{I}{B}=r.
\end{equation}
Substituting the simplification in \eqref{lemma_RAPS_dec_30} into \eqref{lemma_RAPS_dec_20} follows the claim in \eqref{lemma_RAPS_dec_claim_1}. To prove the claim in \eqref{lemma_RAPS_dec_claim_2} we can use the same argument that we used in proving \eqref{lemma_RAPS_dec_claim_1} to show that 
\begin{equation}\label{lemma_RAPS_dec_40}
\mathbb{E}_{\ccalI^t}\!\left[ \|   \bbx_{t+1}\!-\!\bbx^t\|^2\! \mid \ccalF^t   \right]
	\!=\!\frac{{{B-1}\choose{I-1}}}{{{B}\choose{I}}} (\gamma^t)^2\!  \left\|\nabla f( \bbx^{t}, \bbTheta^t)\right\|^2\!\!\!.
\end{equation}
By substituting the simplification in \eqref{lemma_RAPS_dec_30} into \eqref{lemma_RAPS_dec_40} the claim in \eqref{lemma_RAPS_dec_claim_2} follows.

\subsection{Proof of Proposition \ref{martingale_prop}}\label{apx_martingale_prop}


By considering the Taylor's expansion of $F(\bbx^{t+1})$ near the point $\bbx^t$ and observing the Lipschitz continuity of gradients $\nabla F$ with constant $M$ we obtain that the average objective function $F(\bbx^{t+1}) $  is bounded above by
\begin{equation}\label{martingale_10}
F(\bbx^{t+1})\leq F(\bbx^{t}) +\nabla F(\bbx^{t})^T (\bbx^{t+1}-\bbx^{t})+\frac{M}{2}\|\bbx^{t+1}-\bbx^{t}\|^2.
\end{equation}
Compute the expectation of the both sides of \eqref{martingale_10} with respect to the random set $\ccalI^t$ given the observed set of information $\ccalF^t$. Substitute $\mathbb{E}_{\ccalI^t}\!\left[{\bbx^{t+1}-\bbx^{t}\mid \ccalF^t}\right] $ and $\mathbb{E}_{\ccalI^t}\!\left[\|{\bbx^{t+1}-\bbx^{t}\|^2\mid \ccalF^t}\right] $ with their simplifications in \eqref{lemma_RAPS_dec_claim_1} and \eqref{lemma_RAPS_dec_claim_2}, respectively, to write
\begin{align}\label{martingale_20}
\mathbb{E}_{\ccalI^t}\left[F(\bbx^{t+1})\mid \ccalF^t\right]
	&\leq 
	 F(\bbx^{t}) - {r\gamma^t }{}\ \nabla F(\bbx^{t})^T\nabla f( \bbx^{t}, \bbTheta^t)\nonumber\\
	 &\quad+ \frac{rM(\gamma^t)^2}{2 }\ \left\|\nabla f( \bbx^{t}, \bbTheta^t)\right\|^2.
\end{align}
Notice that the stochastic gradient $\nabla f( \bbx^{t}, \bbTheta^t)$ is an unbiased estimate of the average function gradient $ \nabla F(\bbx^{t})$. Therefore, we obtain $\mathbb{E}_{\bbTheta^t} \left[ \nabla f( \bbx^{t}, \bbTheta^t) \mid \ccalF^t\right]= \nabla F(\bbx^{t})$. Observing this relation and considering the assumption in \eqref{ekhtelaf}, the expected value of \eqref{martingale_20} with respect to the set of realizations $\bbTheta^t$ can be written as
\begin{align}\label{martingale_30}
\mathbb{E}_{\ccalI^t,\bbTheta^t}\left[F(\bbx^{t+1})\mid \ccalF^t\right]
	&\leq 
	 F(\bbx^{t}) - {r\gamma^t}{}\ \left\|\nabla F(\bbx^{t})\right\|^2
	 \nonumber\\
	 &\quad + \frac{rM(\gamma^t)^2 K}{2 }.
\end{align}
Subtracting the optimal objective function value $F(\bbx^*)$ form the both sides of \eqref{martingale_30} implies that
\begin{align}\label{martingale_40}
&\mathbb{E}_{\ccalI^t,\bbTheta^t}\left[F(\bbx^{t+1})-F(\bbx^*)\mid \ccalF^t\right]
		\\
	 &\quad
	 \leq 
	 F(\bbx^{t}) -F(\bbx^*)
	 - r\gamma^t \ \left\|\nabla F(\bbx^{t})\right\|^2
	 + \frac{rM(\gamma^t)^2 K}{2 }. \nonumber
\end{align}
We proceed to find a lower bound for the gradient norm $\| \nabla F(\bbx^{t})\|$ in terms of the objective value error $F(\bbx^{t}) -\ F(\bbx^*)$. Assumption \ref{convexity_assumption} states that the average objective function $F$ is strongly convex with constant $m>0$. Therefore, for any $\bby,\bbz\in \reals^p$ we can write
\begin{equation}\label{martingale_50}
   F(\bby) \geq\ F(\bbz) +\nabla F(\bbz)^{T}(\bby-\bbz)   
   + {{m}\over{2}}\|{\bby - \bbz}\|^{2}.
\end{equation}
For fixed $\bbz$, the right hand side of \eqref{martingale_50} is a quadratic function of $\bby$ whose minimum argument we can find by setting its gradient to zero. Doing this yields the minimizing argument $\hby = \bbz- (1/m) \nabla  F(\bbz)$ implying that for all $\bby$ we must have
\begin{alignat}{2}\label{martingale_60}
F(\bby) \geq\ 
    &\ F(\bbw) +\nabla F(\bbz)^{T}(\hby-\bbz)   
   + {{m}\over{2}}\|{\hby - \bbz}\|^{2} \nonumber \\
   \ =\ 
         &\ F(\bbz) - \frac{1}{2m} \| \nabla F(\bbz)\|^{2} .
\end{alignat}
Observe that the bound in \eqref{martingale_60} holds true for all $\bby$ and $\bbz$. Setting values $\bby=\bbx^{*}$ and $\bbz=\bbx^{t}$  in \eqref{martingale_60} and rearranging the terms yields a lower bound for the squared gradient norm $\|\nabla F(\bbx^t)\|^2$ as
\begin{equation}\label{martingale_70}
 \| \nabla F(\bbx^{t})\|^{2} \geq 2m( F(\bbx^{t}) -   F(\bbx^*) ).
\end{equation} 
Substituting the lower bound in \eqref{martingale_70} by the norm of gradient square $ \| \nabla F(\bbx^{t})\|^{2}$ in \eqref{martingale_40} follows the claim in \eqref{martingale_prop_claim}.

\subsection{Proof of Theorem \ref{RAPSA_convg_thm}}\label{apx_RAPSA_convg_thm}


We use the relationship in \eqref{martingale_prop_claim} to build a supermartingale sequence. To do so, define the stochastic process $\alpha^t$ as
\begin{equation}\label{martingale_41}
   \alpha^t := F(\bbx^{t})-F(\bbx^*) +\frac{rMK}{2} \sum_{u=t}^{\infty}  (\gamma^u)^2 .
\end{equation}
Note that $\alpha^t$ is well-defined because $\sum_{u=t}^{\infty}  (\gamma^u)^2 \leq\sum_{u=0}^{\infty}  (\gamma^u)^2 <\infty$ is summable. Further define the sequence $\beta_t$ with values
\begin{equation}\label{martingale_42}
   \beta^t :=\ {2m\gamma^t r} (F(\bbx^{t})-F(\bbx^*) ).
\end{equation}
The definitions of sequences $\alpha^t$ and $\beta^t$ in \eqref{martingale_41} and \eqref{martingale_42}, respectively, and the inequality in \eqref{martingale_prop_claim} imply that the expected value $\alpha^{t+1}$ given $\ccalF^t$ can be written as
\begin{equation}\label{martingale_43}
   \E{\alpha^{t+1} \given \ccalF^t} \ \leq\ \alpha^t - \beta^t.
\end{equation}
Since the sequences $\alpha^t$ and $\beta^t$ are nonnegative it follows from \eqref{martingale_43} that they satisfy the conditions of the supermartingale convergence theorem. 
Therefore, we obtain that: (i) The sequence $\alpha^t$ converges almost surely to a limit. (ii) The sum $\sum_{t=0}^{\infty}\beta^t < \infty$ is almost surely finite. The latter result yields
\begin{equation}\label{martingale_44}
   \sum_{t=0}^{\infty} {2m\gamma^t r} (F(\bbx^{t})-F(\bbx^*) ) < \infty.
       \qquad\text{a.s.}
\end{equation}
Since the sequence of step sizes is non-summable there exits a subsequence of sequence $F(\bbx^{t})-F(\bbx^*)$ which is converging to null. This observation is equivalent to almost sure convergence of $\liminf F(\bbx^{t})-F(\bbx^*)$ to null
\begin{equation}\label{martingale_45}
   \liminf_{t\to  \infty} F(\bbx^{t})-F(\bbx^*)=0.
       \qquad\text{a.s.}
\end{equation}
Based on the martingale convergence theorem for the sequences $\alpha^t$ and $\beta^t$ in relation \eqref{martingale_43}, the sequence $\alpha^t$ almost surely converges to a limit. Consider the definition of $\alpha^t$ in \eqref{martingale_41}. Observe that the sum $\sum_{u=t}^{\infty}  (\gamma^u)^2$ is deterministic and its limit is null. Therefore, the sequence of the objective function value error $F(\bbx^t)-F(\bbx^*)$ almost surely converges to a limit. This observation in association with the result in \eqref{martingale_45} implies that the whole sequence of $F(\bbx^t)-F(\bbx^*)$ converges almost surely to null,  
\begin{equation}\label{martingale_71}
   \lim_{t\to  \infty}\ F(\bbx^{t}) -   F(\bbx^*) =0.
       \qquad\text{a.s.}
\end{equation}
The last step is to prove almost sure convergence of the sequence $\|\bbx^t-\bbx^*\|^2$ to null, as a result of the limit in \eqref{martingale_71}. To do so, we follow by proving a lower bound for the objective function value error $F(\bbx^{t}) -   F(\bbx^*)$ in terms of the squared norm error $\|\bbx^t-\bbx^*\|^2$.
According to the strong convexity assumption, we can write the following inequality 
\begin{equation}\label{martingale_72}
F(\bbx^t)\geq F(\bbx^*)+\nabla F(\bbx^*)^T(\bbx^t-\bbx^*)+\frac{m}{2}\|\bbx^t-\bbx^*\|^2.
\end{equation}
Observe that the gradient of the optimal point is the null vector, i.e., $\nabla F(\bbx^*)=\bb0$. This observation and rearranging the terms in \eqref{martingale_72} imply that 
\begin{equation}\label{martingale_73}
F(\bbx^t) - F(\bbx^*)\geq \frac{m}{2}\|\bbx^t-\bbx^*\|^2.
\end{equation}
The upper bound in \eqref{martingale_73} for the squared norm $\|\bbx^t-\bbx^*\|^2$ in association with the fact that the sequence $F(\bbx^t) - F(\bbx^*)$ almost surely converges to null, leads to the conclusion that the sequence $\|\bbx^t-\bbx^*\|^2$ almost surely converges to zero. Hence, the claim in \eqref{rapsa_as_convg} is valid.

The next step is to study the convergence rate of RAPSA in expectation. In this step we assume that the diminishing stepsize is defined as $\gamma^t=\gamma^0 T^0/(t+T^0)$. Recall the inequality in \eqref{martingale_prop_claim}. Substitute $\gamma^t$ by $\gamma^0 T^0/(t+T^0)$ and compute the expected value of \eqref{martingale_prop_claim} given $\ccalF^0$ to obtain
\begin{align}\label{martingale_90}
&\E{F(\bbx^{t+1})-F(\bbx^*)}\\
& 
	\leq 
	\!\left( 1- \frac{2mr \gamma^0 T^0}{(t+T^0)} \right)\E{F(\bbx^{t}) -F(\bbx^*)}+ \frac{rMK(\gamma^0 T^0)^2}{2 (t+T^0)^2}.\nonumber
\end{align}
We use the following lemma to show that the result in \eqref{martingale_90} implies  sublinear convergence of the sequence of expected objective value error $\E{F(\bbx^{t})-F(\bbx^*)}$.

%
\begin{lemma}\label{lecce22}
Let $c>1$, $b>0$ and $t^0 > 0$ be given constants and $u_{t}\geq 0$ be a nonnegative sequence that satisfies 
\begin{equation}\label{claim23}
   u^{t+1} \leq \left( 1- \frac{c}{t+t^0} \right) u^{t} 
                 + \frac{b}{{(t+t^0)}^{2}}\ ,
\end{equation}
{for all times $t\geq0$}. The sequence $u^t$ is then bounded as
\begin{equation}\label{lemma3_claim}
u^{t} \leq\  \frac{Q}{t+t^{0}},
\end{equation}
for all times $t\geq0$, where the constant $Q$ is defined as $ Q:=\max \{{b}/({c-1}),\ t^{0} u^{0} \}$ .
\end{lemma}
\begin{myproof}
See \cite{Nemirovski}.
\end{myproof}

Lemma \ref{lecce22} shows that if a sequence  $u^t$ satisfies the condition in \eqref{claim23} then the sequence $u^t$ converges to null at least with the rate of $O(1/t)$. By assigning values $t^0=T^0$, $u^t=\E{F(\bbx^t)-F(\bbx^*)}$, $c=2m r\gamma^0 T^0$, and $b=rM  K(\gamma^0 T^0)^2/2$, the relation in \eqref{martingale_90} implies that the inequality in \eqref{claim23} is satisfied for the case that $2mr\gamma^0 T^0>1$. Therefore, the result in \eqref{lemma3_claim} holds and we can conclude that 
\begin{equation}\label{martingale_100}
\E{F(\bbx^t)-F(\bbx^*)}\leq \frac{ C}{t+T^0},
\end{equation}
where the constant $C$ is defined as 
\begin{equation}\label{martingale_110}
C= \max\left\{\frac{rMK (\gamma^0 T^0)^2}{4rm\gamma^0 T^0-2},\ T^0(F(\bbx^0)-F(\bbx^*))\right\}.
\end{equation}

\subsection{Proof of Theorem \ref{RAPSA_convg_thm_finite}}\label{apx_RAPSA_convg_thm_finite}


To prove the claim in \eqref{rapsa_as_convg_finite} we use the relationship in \eqref{martingale_prop_claim} to construct a supermartingale. Define the stochastic process $\alpha^t$ with values 
\begin{equation}\label{finite_10}
\alpha^t\!:=\!\left( F(\bbx^t)-F(\bbx^*) \right) \times
	 \mathbf{1}\!\left\{\min_{u\leq t}  F(\bbx^u)-F(\bbx^*) \!>\! \frac{\gamma M K}{4m}\right\}
\end{equation}
The process $\alpha^t$ tracks the optimality gap $F(\bbx^t)-F(\bbx^*)$ until the gap becomes smaller than $\gamma M K/{2m}$ for the first time at which point it becomes $\alpha^t=0$. Notice that the stochastic process $\alpha^t$ is always non-negative, i.e., $\alpha^t\geq0$. Likewise, we define the stochastic process $\beta^t$ as
\begin{align}\label{finite_20}
\beta^t&:={2\gamma m r}{}\left( F(\bbx^t)-F(\bbx^*)-\frac{\gamma M K}{4m} \right) 
\nonumber\\
&\qquad
\times
	 \mathbf{1}\left\{\min_{u\leq t}\  F(\bbx^u)-F(\bbx^*) > \frac{\gamma M K}{4m}\right\},
\end{align}
which follows $2\gamma m r\left( F(\bbx^t)-F(\bbx^*)-{\gamma M K}/{4m} \right) $ until the time that the optimality gap $F(\bbx^t)-F(\bbx^*)$ becomes smaller than $\gamma M K/{2m}$ for the first time. After this moment the stochastic process $\beta^t$ becomes null. According to the definition of $\beta^t$ in \eqref{finite_20}, the stochastic process satisfies $\beta^t\geq0$ for all $t\geq0$. Based on the relationship \eqref{martingale_prop_claim} and the definitions of stochastic processes $\alpha^t$ and $\beta^t$ in \eqref{finite_10} and \eqref{finite_20} we obtain that for all times $t\geq0$
\begin{equation}\label{finite_30}
\E{\alpha^{t+1} \mid \ccalF^t} \leq \alpha^t-\beta^t.
\end{equation}
To check the validity of \eqref{finite_30} we first consider the case that $\min_{u\leq t}\  F(\bbx^u)-F(\bbx^*) > {\gamma M K}{/4m}$ holds. In this scenario we can simply the stochastic processes in \eqref{finite_10} and \eqref{finite_20} as $\alpha^t=F(\bbx^t)-F(\bbx^*)$ and $\beta^t=2\gamma m r\left( F(\bbx^t)-F(\bbx^*)-{\gamma M K}/{4m} \right)$. Therefore, according to the inequality in \eqref{martingale_prop_claim} the result in \eqref{finite_30} is valid. The second scenario that we check is $\min_{u\leq t}\  F(\bbx^u)-F(\bbx^*) \leq {\gamma M K}{/4m}$. Based on the definitions of stochastic processes $\alpha^t$ and $\beta^t$, both of these two sequences are equal to 0. Further, notice that when $\alpha^t=0$, it follows that $\alpha^{t+1}=0$. Hence, the relationship in \eqref{finite_30} is true.

Given the relation in \eqref{finite_30} and non-negativity of stochastic processes $\alpha^t$ and $\beta^t$ we obtain that $\alpha^t$ is a supermartingale. The supermartingale convergence theorem yields: i) The sequence $\alpha^t$ converges to a limit almost surely. ii) The sum $\sum_{t=1}^\infty \beta^t$ is finite almost surely. The latter result implies that the sequence $\beta^t$ is converging to null almost surely. I.e., 
\begin{equation}\label{finite_40}
\lim_{t \to \infty} \beta^t \ =\ 0 \quad  a.s. 
\end{equation}
Based on the definition of $\beta^t$ in \eqref{finite_20}, the limit in \eqref{finite_40} is true if one of the following events holds: i) The indicator function is null after for large $t$. ii) The limit $\lim_{t\to \infty} \left( F(\bbx^t)-F(\bbx^*)-{\gamma M K}/{4m} \right) =0$ holds true. From any of these two events we it is implied that  
\begin{equation}\label{finite_50}
\liminf_{t\to \infty}\ F(\bbx^t)-F(\bbx^*)\ \leq \ \frac{\gamma M K}{4m}\quad a.s.
\end{equation}
Therefore, the claim in \eqref{rapsa_as_convg_finite} is valid. The result in \eqref{finite_50} shows the objective function value sequence $F(\bbx^t)$ almost sure converges to a neighborhood of the optimal objective function value $F(\bbx^*)$. 

We proceed to prove the result in \eqref{rapsa_rate_finite}. Compute the expected value of \eqref{martingale_prop_claim}  given $\ccalF^0$ and set $\gamma^t=\gamma$ to obtain
\begin{align}\label{finite_constant_10}
\E{F(\bbx^{t+1})-F(\bbx^*)}
	&\leq 
	\left( 1- 2m  \gamma r \right)\E{ F(\bbx^{t}) -F(\bbx^*)}
	\nonumber\\
	&\qquad
	+ \frac{rM K\gamma^2}{2 }.
\end{align}
Notice that the expression in \eqref{finite_constant_10} provides an upper bound for the expected value of objective function error $\E{F(\bbx^{t+1})-F(\bbx^*)}$ in terms of its previous value $\E{F(\bbx^{t})-F(\bbx^*)}$ and an error term. Rewriting the relation in \eqref{finite_constant_10} for step $t-1$ leads to 
\begin{align}\label{finite_constant_20}
\E{F(\bbx^{t})-F(\bbx^*)}
	&\leq 
	\left( 1- 2m  \gamma r \right)\E{ F(\bbx^{t-1}) -F(\bbx^*)}
		\nonumber\\
	&\qquad
	+ \frac{rM K\gamma^2}{2 }.
\end{align}
Substituting the upper bound in \eqref{finite_constant_20} for the expectation $\E{F(\bbx^{t})-F(\bbx^*)}$ in \eqref{finite_constant_10} follows an upper bound for the expected error $\E{F(\bbx^{t+1})-F(\bbx^*)}$ as
\begin{align}\label{finite_constant_30}
\E{F(\bbx^{t+1})\!-\!F(\bbx^*)}
	&\!\leq\! 
	\left( 1- 2m  \gamma r \right)^2\E{ F(\bbx^{t-1}) \!-\!F(\bbx^*)}
		\nonumber\\
	&\!\quad\!
	+ \frac{rM K\gamma^2}{2 }\left(1+\left( 1\!- \!2mr \gamma\right)\right)\!.
\end{align}
By recursively applying the steps in \eqref{finite_constant_20}-\eqref{finite_constant_30} we can bound the expected objective function error $\E{F(\bbx^{t+1})-F(\bbx^*)}$ in terms of the initial objective function error $F(\bbx^{0}) -F(\bbx^*)$ and the accumulation of the errors as
\begin{align}\label{finite_constant_40}
\E{F(\bbx^{t+1})\!-\!F(\bbx^*)}
	&\leq 
	\left( 1- 2m  \gamma r \right)^{t+1}( F(\bbx^{0}) -F(\bbx^*))
	\nonumber\\
	&\quad
	+ \frac{rM K\gamma^2}{2 }
	\sum_{u=0}^t
	\left( 1- {2mr\gamma}{} \right)^u\!\!\!.
\end{align}
Substituting $t$ by $t-1$ and simplifying the sum in the right hand side of \eqref{finite_constant_40} yields 
\begin{align}\label{finite_constant_50}
\E{F(\bbx^{t})-F(\bbx^*)}
	&\leq 
	\left( 1- 2m  \gamma r \right)^{t}( F(\bbx^{0}) -F(\bbx^*))
	\nonumber\\
	&\quad
	+ \frac{M  K\gamma}{4 m}
	\left[ 1- \left( 1- {2mr \gamma}{} \right)^t\right].
\end{align}
Observing that the term $1- \left( 1- {2mr \gamma}{} \right)^t$ in the right hand side of \eqref{finite_constant_50} is strictly smaller than $1$ for the stepsize $\gamma<1/(2mr)$, the claim in \eqref{rapsa_rate_finite} follows.